\documentclass[10pt, a4paper]{article}

\usepackage[final]{lrec2026} 
\usepackage{booktabs}
\usepackage{tcolorbox}
\tcbuselibrary{listings, skins, breakable}
\usepackage{tabularx}
\usepackage{adjustbox}
\tcbuselibrary{breakable}
\usepackage{graphicx}
\usepackage{caption}
\usepackage{subcaption}
\usepackage{float}
\usepackage{microtype}

\title{Modeling Topics and Sociolinguistic Variation in Code-Switched Discourse: Insights from Spanish-English and Spanish-Guaraní}

\name{Nemika Tyagi$^{1*}$\thanks{$^{*}$Main and Corresponding Author.}, Nelvin Licona Guevara$^{1}$, Olga Kellert$^{1}$}

\address{$^{1}$Arizona State University, USA\\
         \{ntyagi8, nliconag, olga.kellert\}@asu.edu\\}

\abstract{
This study presents an LLM-assisted annotation pipeline for the sociolinguistic and topical analysis of bilingual discourse in two typologically distinct contexts: Spanish-English and Spanish-Guaraní. Using large language models, we automatically labeled topic, genre, and discourse-pragmatic functions across a total of 3,691 code-switched sentences, integrated demographic metadata from the \textit{Miami Bilingual Corpus}, and enriched the \textit{Spanish-Guaraní} dataset with new topic annotations. The resulting distributions reveal systematic links between gender, language dominance, and discourse function in the Miami data, and a clear diglossic division between formal Guaraní and informal Spanish in Paraguayan texts. These findings replicate and extend earlier interactional and sociolinguistic observations with corpus-scale quantitative evidence. The study demonstrates that large language models can reliably recover interpretable sociolinguistic patterns traditionally accessible only through manual annotation, advancing computational methods for cross-linguistic and low-resource bilingual research. 
 \\ \newline  }

\begin{document}

\maketitleabstract

\section{Introduction}

\noindent
Code-switching is the alternation between two or more languages within a single discourse and it is a pervasive feature of bilingual and multilingual communication. It plays a central role in how speakers negotiate identity, stance, and alignment in everyday interaction. Far from being random, language switching is systematically tied to discourse and pragmatic functions.  Early interactional work (\citealt{Gumperz1982}; \citealt{Auer1998}) demonstrated that switches can index conversational frames, signal quotations or reported speech, and mark topic or participant shifts. Subsequent studies expanded this view by showing that switching contributes to narrative organization, topic management, and information flow (\citealt{Bullock2009}; \citealt{Fricke2016}; \citealt{Kootstra2020}; \citealt{GardnerChloros2009}). Pragmatic and cognitive approaches have also emphasized that code-switching reflects fine-grained speaker control over audience design, style, and footing (\citealt{gafaranga2011}; \citealt{MyersScotton1993}; \citealt{Matras2009}). Yet, these rich qualitative insights rely on manual annotation and small datasets, limiting their scalability and cross-linguistic coverage.

Computational linguists have sought to automate the detection and structural modeling of code-switching. Early NLP studies focused on identifying switch points or predicting the matrix language using surface and syntactic features (\citealt{Solorio2008a}; \citealt{Solorio2008b}; \citealt{Jamatia2015}; \citealt{Molina2016}). More recent work employs neural architectures to improve token-level language identification and sequence tagging (\citealt{Winata2018}; \citealt{Bhat2023}), and shared tasks have promoted benchmark datasets across language pairs (\citealt{Molina2016}; \citealt{Patwa2020}). However, these methods primarily model \textit{where} switches occur rather than \textit{why}. Few computational studies explicitly represent the discourse, pragmatic, or topical motivations underlying switching (\citealt{Liu2021}; \citealt{Zirn2023}). Consequently, quantitative generalizations about the communicative functions of switching remain scarce. Sociolinguistic variation adds further complexity. Classic studies link code-switching to speaker demographics, community norms, and social meaning (\citealt{Poplack1980}; \citealt{Romaine1995}; \citealt{Toribio2004}; \citealt{Gumperz1982}). Yet, most available corpora lack demographic annotation, hindering systematic sociolinguistic analysis. Low-resource and endangered languages are especially underrepresented (\citealt{Joshi2020}; \citealt{Ponti2020}), creating a skewed empirical base that privileges shallow studies on the nature of code-switching for only high-resource bilingual pairs such as English-Spanish or Mandarin-English. 

Large language models (LLMs) now present an opportunity to bridge linguistic and computational perspectives. 
Their ability to model contextual and cross-lingual semantics allows for new forms of discourse-level analysis (\citealt{Ruder2019}; \citealt{Kirk2023}). However, even the most advanced multilingual models still struggle with mixed-language input (\citealt{Zhang2023}; \citealt{Potter2024}; \citealt{Cahyawijaya2021}). Such errors arise because multilingual models are trained primarily on monolingual or parallel text, leaving conversational and dynamically switched data understudied. 

\noindent
This study addresses this gap by exploring how topic modelling, pragmatic function, and speaker attributes shape bilingual discourse. We integrate LLM-based topic modeling with discourse-pragmatic annotation to examine \textit{why} switches occur in addition to \textit{where}. By focusing on both Spanish-English and Spanish-Guaraní, we extend typological and regional coverage beyond high-resource language pairs and include underrepresented  bilingual contexts. This approach unites humanistic and computational perspectives, providing scalable yet interpretable analyses of bilingual communication.

\noindent
Our main contributions are as follows:
\vspace{-\topsep}
\begin{itemize}
    \item Propose a GPT-based topic classification pipeline with interpretable, linguistically grounded category refinement.
    \item Integrate sociolinguistic metadata (gender, age, language dominance) with bilingual datasets to enable demographic and cross-linguistic comparison.
    \item Release enhanced bilingual corpora annotated for topics and discourse functions, with visualization of switch and topic distributions\footnote{Data is available at \url{https://github.com/N3mika/topicmodelling}.}.
    \item Provide comparative insights into discourse and social variables across high- and low-resource bilingual contexts.
\end{itemize}

\noindent In doing so, our study moves toward a more inclusive and linguistically grounded understanding of bilingual communication, highlighting how computational methods can capture the social and discourse complexity of multilingual speakers.

\section{Background and Related Work}

\paragraph{Discourse and Pragmatic Perspectives}
Foundational interactional studies positioned code-switching as a discourse-organizational and identity-constructing practice, linking language alternation to topic shifts, quotations, and stance (\citealt{Gumperz1982}; \citealt{Auer1998}; \citealt{Bullock2009}). Later pragmatic and cognitive accounts further showed that switches contribute to affective alignment, audience design, and narrative coherence (\citealt{MyersScotton1993}; \citealt{Matras2009}; \citealt{Fricke2016}). Despite these insights, such analyses relied on manual coding, limiting systematic quantification of discourse functions across corpora. Sociolinguistic research emphasizes that code-switching varies with gender, age, and social networks (\citealt{Poplack1980}; \citealt{Romaine1995}; \citealt{Toribio2004}; \citealt{Sankoff1998}). However, most publicly available corpora lack demographic metadata and remain skewed toward high-resource bilingual pairs (\citealt{Joshi2020}; \citealt{Ponti2020}). Low-resource languages are particularly underrepresented, restricting comparative and inclusive analysis.

\paragraph{Computational Modeling of Code-Switching}
Early computational work focused on token-level language identification and matrix language prediction (\citealt{Solorio2008a}; \citealt{Molina2016}; \citealt{Patwa2020}). Neural architectures have since improved language tagging accuracy (\citealt{Winata2018}; \citealt{Bhat2023}), yet most systems do not incorporate pragmatic or topical cues. 
A few recent studies have begun exploring discourse motivations and contextual embeddings (\citealt{Liu2021}; \citealt{Zirn2023}), but these approaches are still limited in scope and coverage. Large language models (LLMs) offer new possibilities for discourse-level multilingual analysis (\citealt{Ruder2019}; \citealt{Kirk2023}). Nonetheless, empirical evaluations show that multilingual models often misinterpret mixed-language input, treating code-switching as noise rather than strategic discourse choice (\citealt{Zhang2023}; \citealt{Potter2024}; \citealt{Cahyawijaya2021}). These findings suggest the need for linguistically grounded evaluation resources that capture both structural and sociolinguistic aspects of bilingual communication. Our work integrates these strands by combining LLM-based topic modeling with discourse-pragmatic annotation and sociolinguistic metadata, linking functional interpretations of switching with scalable computational methods.

\section{Experiments}
\subsection{Datasets}

\paragraph{Miami Corpus (English-Spanish).}
The Miami corpus (\citetlanguageresource{BangorMiami2014}) is part of the BilingBank repository and was collected between 2008 and 2011 by the ESRC Centre for Research on Bilingualism, Bangor University.  
It consists of transcripts of informal conversations among 84 bilingual speakers in Miami, USA.  
Recordings capture spontaneous interactions, later transcribed and pseudonymized. 
Participants provided demographic information through post-recording questionnaires, enabling sociolinguistic analysis.  
For the present study, we extract the subset of 2,825 sentences containing intra-sentential code-switching between English and Spanish.

\paragraph{GUA-SPA Corpus (Spanish-Guaraní).}
The Guaraní dataset originates from the GUA-SPA shared task at IberLEF 2023 (\citetlanguageresource{GUA-SPA2023}), which contains Guaraní-Spanish mixed texts sourced from Paraguayan tweets and news articles.  
The full corpus comprises 1,500 texts and about 25k tokens. 
We use the training-set portion and further extract a subset containing sentences with intra-sentential switching, represented at the token level.
All Spanish variants and named-entity tags were merged under \textit{spa}, while foreign and unclassified tokens such as punctuations and emojis were grouped as \textit{other}.  

\paragraph{Subset Statistics.}
Table~\ref{tab:corpus-stats} summarizes the statistics of the two code-switched datasets and their corresponding subsets used in this study, including their token counts, language proportions, and average code-switching density measured as adjacent language changes per sentence.

\begin{table}[t]
\centering
\footnotesize
\begin{tabular}{lrrr}
\toprule
Corpus & Sentences & Tokens & Avg token/sent. \\
\midrule
Miami & 2825 & 29.7k & 10.5 \\
Spa-Gua & 866 & 15.6k & 18.0 \\
\bottomrule
\end{tabular}

\vspace{0.3em}
\begin{tabular}{l@{\hspace{1em}}l}
\toprule
Corpus & Lang.\ proportion (\%) \\
\midrule
Miami & spa~48.4; eng~40.1; punc~9.4; eng\&spa~2 \\
Spa-Gua & spa~42.6; gn~38.7; other~16.6; gn\&spa~2.1 \\
\bottomrule
\end{tabular}
\caption{Summary of code-switched subsets and token-level language proportions. 
Language tags: \textit{spa} = Spanish, \textit{eng} = English, 
\textit{gn} = Guaraní, \textit{punc} = punctuation, 
\textit{other} = punctuation, special characters, or emojis, 
\textit{eng\&spa}/\textit{gn\&spa} = ambiguous mixed-language tokens.}
\label{tab:corpus-stats}
\end{table}

\subsection{Experimental Setup}
The annotated datasets were created by inferencing the \texttt{gpt-4.1-2025-04-14} model through the OpenAI API. Each sentence, together with speaker and situational metadata, was processed individually using deterministic parameters (\texttt{temp}=0, \texttt{max\_tokens}=200). The pipeline constructed structured prompts containing sentence ID, language tag, and contextual information, and normalized the model’s outputs to canonical topic and function labels. Annotation was conducted in batches of 50-100 sentences, covering 2{,}825 code-switched sentences from the Miami corpus and 866 from the Spanish-Guaraní dataset (see Table~\ref{tab:corpus-stats}).

\section{Methods}

\subsection{Category Selection}
We began by sampling 30 random sentences from each dataset and collaboratively developed an initial categorization schema with two bilingual annotators.  
For the Miami corpus, two dimensions were defined: \textit{Topics} (domain or content areas) and \textit{Functions} (discourse or pragmatic roles).  
For the Spanish--Guaraní dataset, three dimensions were introduced: \textit{Formality}, \textit{Genre}, and \textit{Topic}.  
After initial annotation, the schemas were iteratively refined by reviewing an additional batch of 30 sentences to ensure coverage of the linguistic and contextual diversity of each corpus.  
Semantically overlapping categories were merged, and concise explanatory notes were added to clarify the scope of each category for future annotation consistency.  
All categories were cross-validated by both annotators before pipeline implementation.  
While these taxonomies were created through careful linguistic reasoning, we acknowledge that topic and discourse categorization is inherently subjective; different annotators or frameworks could yield alternative but equally valid interpretations.  
The goal was to establish a practical and interpretable schema for enhancing bilingual corpora and enabling downstream sociolinguistic analysis.

Below are the details of the multi-tiered Annotation Schemas that we developed for the Miami and SPA-GUA corpus.
\begin{tcolorbox}[colback=gray!3!white, colframe=gray!50!black,
    boxrule=0.5pt,
    arc=2pt,
    title=\textbf{Miami Corpus Annotation Schema},
    fonttitle=\bfseries,
    coltitle=white,
    left=3pt,right=3pt,top=3pt,bottom=3pt]

\begin{tcolorbox}[colback=blue!5!white, colframe=blue!25!black, boxrule=0.3pt, arc=1pt, left=3pt, right=3pt, top=3pt, bottom=3pt,
title=\textbf{Functions (choose one primary; secondary allowed if clearly two)}]
\textbf{TechnicalTermInsertion:} inserting domain-specific words or tool names.\\
\textbf{ProperNounNamedEntity:} naming a person, place, brand, or award.\\
\textbf{PrecisionLexicalGap:} switching for precise expression or lexical need.\\
\textbf{DiscourseMarker:} connective or organizing signals (e.g., \textit{you know, so}).\\
\textbf{TopicShift:} marking a new topic or returning to one.\\
\textbf{Narrative:} embedding a story or recounting a past event.\\
\textbf{Quotation:} reproducing or stylizing another’s voice.\\
\textbf{TurnManagement:} backchannels or acknowledgments (\textit{mmhm, yeah}).\\
\textbf{AddresseeShift:} calling attention or changing addressee (\textit{hey Bob}).\\
\textbf{Directive:} giving orders, requests, or imperatives.\\
\textbf{Repair:} rephrasing, searching for a word, or self-correcting.\\
\textbf{Agreement:} affirming or echoing another speaker’s stance.\\
\textbf{StanceEmphasis:} expressing evaluation, certainty, or irony.\\
\textbf{Humor:} jokes, teasing, or playful language.\\
\textbf{SolidarityIdentity:} in-group markers or swearing showing closeness.
\end{tcolorbox}
\end{tcolorbox}
\begin{tcolorbox}[colback=gray!3!white, colframe=gray!50!black,
    boxrule=0.5pt,
    arc=2pt,
    fonttitle=\bfseries,
    coltitle=white,
    left=3pt,right=3pt,top=3pt,bottom=3pt]

\begin{tcolorbox}[colback=blue!5!white, colframe=blue!25!black, boxrule=0.3pt, arc=1pt, left=3pt, right=3pt, top=3pt, bottom=3pt, title=\textbf{Topics (choose one; if mixed, choose the dominant)}]
\textbf{Workplace\_Technical:} technical terms, commissioning, CAD, architecture terms.\\
\textbf{Education\_YouthOrganizations:} school, certificates, scouts, permission slips.\\
\textbf{Architecture\_Design:} materials, styles, famous architects.\\
\textbf{Office\_Logistics:} supplies, scheduling, file paths, emails.\\
\textbf{Narratives\_Quotations:} recounting past events or reported speech.\\
\textbf{Casual\_EverydayTalk:} greetings, jokes, small talk, banter.\\
\textbf{Affect\_Identity:} swearing, nicknames, identity/solidarity markers.\\
\textbf{ProperNouns\_NamedEntities:} sentences dominated by names, places, or awards.
\end{tcolorbox}
\end{tcolorbox}

\vspace{1.5em}

\begin{tcolorbox}[colback=gray!3!white, colframe=gray!50!black,
    boxrule=0.5pt,
    arc=2pt,
    title=\textbf{Spanish-Guaraní Corpus Annotation Schema},
    fonttitle=\bfseries,
    coltitle=white,
    left=3pt,right=3pt,top=3pt,bottom=3pt]
\begin{tcolorbox}[colback=orange!5!white, colframe=orange!40!black, boxrule=0.3pt, arc=1pt,
left=3pt, right=3pt, top=3pt, bottom=3pt, title=\textbf{Formality (choose one)}]
\textbf{Formal:} official or institutional tone; objective or procedural (e.g., announcements, reports, press releases).\\
\textbf{Informal:} conversational, personal, humorous, or emotional tone; includes slang, emojis, or direct address.
\end{tcolorbox}
\end{tcolorbox}
\begin{tcolorbox}[colback=gray!3!white, colframe=gray!50!black,
    boxrule=0.5pt,
    arc=2pt,
    fonttitle=\bfseries,
    coltitle=white,
    left=3pt,right=3pt,top=3pt,bottom=3pt]
\begin{tcolorbox}[colback=orange!5!white, colframe=orange!40!black, boxrule=0.3pt, arc=1pt,
left=3pt, right=3pt, top=3pt, bottom=3pt, title=\textbf{Genre (choose one)}]
\textbf{News:} objective reports or summaries of events.\\
\textbf{Personal:} emotions, reflections, or personal experiences.\\
\textbf{Politics:} mentions politicians, elections, or government affairs.\\
\textbf{Activism\_Protest:} references to mobilizations or calls to action.\\
\textbf{Culture\_Arts:} music, literature, art.\\
\textbf{Education:} covers schools, universities, or reforms.\\
\textbf{Health:} health, medicine, or COVID-19.\\
\textbf{Environment:} ecology, nature, conservation.\\
\textbf{Sports:} athletic events or teams.\\
\textbf{Entertainment:} celebs, humor, pop culture.\\
\textbf{Commercial:} ads, business, or products.\\
\textbf{Announcement:} schedules, program info.\\
\textbf{Opinion:} commentary or evaluation of public issues.\\
\textbf{Other:} fallback for unclear categories.
\end{tcolorbox}
\end{tcolorbox}
\begin{tcolorbox}[colback=gray!3!white, colframe=gray!50!black,
    boxrule=0.5pt,
    arc=2pt,
    fonttitle=\bfseries,
    coltitle=white,
    left=3pt,right=3pt,top=3pt,bottom=3pt]
\begin{tcolorbox}[colback=orange!5!white, colframe=orange!40!black, boxrule=0.3pt, arc=1pt,
left=3pt, right=3pt, top=3pt, bottom=3pt, title=\textbf{Topics (choose one; if two, mark a secondary)}]
\textbf{Government\_Announcement:} official statements from institutions.\\
\textbf{Legislation\_Policy:} mentions laws, regulations, or legislative actions.\\
\textbf{Protest\_Report:} reports describing protests or demonstrations.\\
\textbf{Mobilization\_Call:} calls for strikes, activism.\\
\textbf{Corruption\_Donations\_Procurement:} references of such.\\
\textbf{PublicAdministration\_Changes:} appointments or administrative shifts.\\
\textbf{Procurement\_Licitation:} references to tenders or contract awards.\\
\textbf{Infrastructure\_Contract:} mentions construction or development projects.\\
\textbf{Transport\_PublicSafety:} transportation or safety-related content.\\
\textbf{Agriculture\_Reactivation:} farming or agrarian reform.\\
\textbf{Rural\_Community\_Issues:} rural life or community concerns.\\
\textbf{Indigenous\_CommunityAid:} Indigenous rights or aid programs.\\
\textbf{Education\_Policy\_University:} education reforms or student activism.\\
\textbf{Cultural\_Event\_Festival:} festivals or public celebrations.\\
\textbf{Cultural\_Heritage\_Archive:} heritage preservation or archives.\\
\textbf{Media\_Broadcast\_Notice:} broadcast or program announcements.\\
\textbf{Legal\_Judicial:} courts, rulings, or judicial.\\
\textbf{Crime\_Investigation:} mentions crimes or investigations.\\
\textbf{Health\_COVID:} COVID-19, vaccines, or health effects.\\
\textbf{PublicHealth\_Services:} hospitals or medical access.\\
\textbf{Environment\_NationalParks:} conservation or protected areas.\\
\textbf{Commercial\_Product:} product promotions or corporate content.\\
\textbf{Shopping\_PersonalPurchase:} consumer life or buying habits.\\
\textbf{Personal\_Emotional:} emotional reflections or personal states.\\
\textbf{Humor\_Rant:} jokes, sarcasm, or venting.\\
\textbf{Sports\_Event:} matches, scores, or athletes.\\
\textbf{Entertainment\_Music\_Film:} mentions music, artists, or movies.\\
\textbf{Opinion\_Commentary:} subjective political or social commentary.\\
\textbf{UserMention\_Request\_Response:} direct replies, mentions, or user interactions.\\
\textbf{Other:} unclear or uncategorizable tweets.
\end{tcolorbox}
\end{tcolorbox}

\subsection{Topic Annotation Workflow}
To operationalize the annotation schemas, we designed a structured prompting workflow for automatic labeling using the \texttt{gpt-4.1} model. Each prompt simulated the detailed instructions a human annotator would follow and consisted of three coordinated components: (1) a \textit{system prompt} that defined the annotator’s role (“You are a careful Spanish-English discourse annotator”) and constrained the model to output only a single JSON object; (2) a \textit{base prompt} that described the input structure and labeling procedure, specifying that each sentence accompanied by speaker and situational metadata such as \textit{speaker}, \textit{age}, \textit{gender}, \textit{situation}, and \textit{lang\_tag} must be tagged with exactly one \texttt{topic} and one \texttt{function}, and an optional \texttt{secondary\_function}; and (3) a set of \textit{instruction lists} enumerating the available topic and function labels defined for each dataset. A schematic representation of this workflow is shown below, summarizing the process from data input and prompt construction to annotation and post-processing.
\begin{tcolorbox}[
    colback=white,
    colframe=gray!60!black,
    boxrule=0.5pt,
    arc=2pt,
    title=\textbf{Annotation Pipeline Structure},
    fonttitle=\bfseries,
    left=3pt,right=3pt,top=3pt,bottom=3pt,
    width=\columnwidth,
    fontupper=\footnotesize,
    label={fig:annotation-pipeline},
    listing only
]

\begin{tcolorbox}[colback=blue!5!white, colframe=blue!50!black, boxrule=0.3pt, arc=1pt,
left=3pt,right=3pt,top=3pt,bottom=3pt, boxsep=2pt]
\textbf{System Instructions:}\\
``You are a careful Spanish-English discourse annotator.\\
Given a sentence and short metadata, assign exactly one primary \texttt{topic} and one primary \texttt{function},\\
and optionally a \texttt{secondary\_function} if clearly present.\\
Be conservative: choose the label that best captures the discourse-level purpose of the sentence.\\
Return \textbf{only} strict JSON no extra text or explanations.'' 
\end{tcolorbox}

\begin{tcolorbox}[colback=orange!5!white, colframe=orange!60!black, boxrule=0.3pt, arc=1pt,
left=3pt,right=3pt,top=3pt,bottom=3pt, boxsep=2pt]
\textbf{Prompt:}\\
``Input fields: \textit{sent\_id, filename, speaker, age, gender, situation, lang\_tag, sentence}.\\
Read the sentence and metadata carefully and select the most fitting labels:\\
-- \texttt{topic}: 1 primary domain label\\
-- \texttt{function}: 1 primary discourse/pragmatic label\\
-- \texttt{secondary\_function}: optional, only if an additional pragmatic role is evident.\\
Use exact category strings from the provided instruction lists.'' 
\end{tcolorbox}

\begin{tcolorbox}[colback=green!5!white, colframe=green!50!black, boxrule=0.3pt, arc=1pt,
left=3pt,right=3pt,top=3pt,bottom=3pt, boxsep=2pt]
\textbf{Example query and expected output:}\\
\textit{Sentence:} \texttt{ay ay yo vi los kneepads.}\\
\textit{Metadata:} (Age 63, Gender F, LangTag spa+eng)\\[2pt]
\textit{Expected output:}\\
\{\texttt{"sent\_id": 916, "topic": "Casual\_EverydayTalk", "function": "TechnicalTermInsertion"}\}
\end{tcolorbox}
Overview of the GPT-based annotation pipeline used for discourse-level topic and function labeling.
\end{tcolorbox}

Few-shot exemplars were appended to the base prompt to illustrate correct labeling behavior for short conversational and code-switched sentences, ensuring consistent adherence to the pragmatic scope of the task. The model was instructed to be conservative, choosing the single label that best captured the topical or discourse-level purpose of each utterance. Each response was required to follow strict formatting, which facilitated automated parsing and normalization.

\subsection{Evaluation Metrics}
To evaluate annotation quality, 30 randomly selected sentences from each annotated dataset were reviewed by bilingual linguists.  
Each verifier assessed the plausibility and fit of the assigned labels.  
The Miami corpus achieved 100\% accuracy for \textit{topic} and \textit{function} labels, and 60\% accuracy for \textit{secondary\_function}.  
The Spanish-Guaraní corpus obtained 94.17\% accuracy across its four fields (\textit{Formality}, \textit{Genre}, \textit{Topic}, and \textit{Secondary\_Topic}).  
These results indicate high reliability of the GPT-based labeling pipeline for both corpora, with minor variation in secondary or ambiguous cases.

\section{Results and Discussion}

\subsection{Miami Corpus}

\subsubsection{Sociolinguistic Topic Modelling}
Table~\ref{tab:miami-topics-gender}, \ref{tab:miami-functions-gender} presents the gender-normalized topic and function distributions for the Miami corpus. 
The topic distribution (Table~\ref{tab:miami-topics-gender}) shows the relative proportions of topics across male and female speakers, 
while the function distribution (Table~\ref{tab:miami-functions-gender}) highlights pragmatic differences across genders. Both male and female speakers predominantly engage in \textit{"Casual\_EverydayTalk"} (60\%), reflecting the conversational and informal character of the recordings. Narrative and quotation contexts are the next most frequent, followed by work-related and technical discussions. Minor yet notable differences emerge: female speakers contribute proportionally more to \textit{"Office\_Logistics"} and \textit{"Education\_YouthOrganizations"}, while male speakers show slightly higher proportions of named-entity references. Pragmatically, both genders favor \textit{"Narrative"} functions, though women exhibit marginally higher rates of directive and solidarity-related functions, consistent with earlier sociolinguistic observations that female bilinguals employ switching for interpersonal alignment (cf. \citealt{Poplack1980}; \citealt{Romaine1995}). The current large-scale annotation provides quantitative support for these qualitative observations across nearly 3,000 bilingual utterances.

\begin{table}[t]
\centering
\scriptsize
\setlength{\tabcolsep}{5.5pt}
\begin{adjustbox}{max width=\columnwidth}
\begin{tabularx}{\linewidth}{@{}lrrr@{}}
\toprule
\textbf{Topic} & \textbf{Men (\%)} & \textbf{Women (\%)} & \textbf{Tot. ($n$)} \\
\midrule
Casual\_EverydayTalk          & 59.8 & 60.1 & 1694 \\
Narratives\_Quotations        & 20.5 & 18.5 &  536 \\
Workplace\_Technical          &  4.8 &  4.8 &  135 \\
Office\_Logistics             &  1.6 &  5.7 &  130 \\
ProperNouns\_NamedEntities    &  7.0 &  3.7 &  130 \\
Education\_YouthOrganizations &  2.5 &  3.4 &   90 \\
Affect\_Identity              &  2.8 &  3.3 &   89 \\
Architecture\_Design          &  1.1 &  0.5 &   18 \\
\midrule
Total Sentences                        & 757  & 2065  & 2822 \\
\bottomrule
\end{tabularx}
\end{adjustbox}
\caption{Topic distribution by gender (normalized by gender totals) in the Miami corpus. Percentages are normalized within each gender.}
\label{tab:miami-topics-gender}
\end{table}

\begin{table}[t]
\centering
\scriptsize
\setlength{\tabcolsep}{8pt}
\begin{adjustbox}{max width=\columnwidth}
\begin{tabularx}{\linewidth}{@{}lrrr@{}}
\toprule
\textbf{Function} & \textbf{Men (\%)} & \textbf{Women (\%)} & \textbf{Total ($n$)} \\
\midrule
PrecisionLexicalGap   & 24.3 & 28.1 & 765 \\
Narrative             & 19.6 & 19.8 & 556 \\
DiscourseMarker       & 12.0 & 12.4 & 348 \\
TechnicalTermInsertion & 10.6 & 10.5 & 296 \\
StanceEmphasis        &  6.9 &  6.1 & 178 \\
ProperNounNamedEntity &  8.5 &  4.5 & 156 \\
Directive             &  4.0 &  6.0 & 153 \\
SolidarityIdentity    &  2.4 &  3.5 &  90 \\
Repair                &  3.4 &  2.3 &  73 \\
Quotation             &  3.3 &  2.2 &  70 \\
TurnManagement        &  2.4 &  1.6 &  52 \\
Agreement             &  1.3 &  1.1 &  33 \\
AddresseeShift        &  0.7 &  1.2 &  30 \\
Humor                 &  0.7 &  0.2 &  10 \\
TopicShift            &  0.1 &  0.4 &  10 \\
UNKNOWN\_FUNCTION     &  0.0 &  0.1 &   2 \\
\midrule
Total Sentences                & 757  & 2065  & 2822 \\
\bottomrule
\end{tabularx}
\end{adjustbox}
\caption{Function distribution by gender (normalized by gender totals) in the Miami corpus. Percentages are normalized within each gender.}
\label{tab:miami-functions-gender}
\end{table}

\subsubsection{Bilingual Asymmetries}
\begin{figure}[t]
  \centering
  \begin{subfigure}[t]{\columnwidth}
    \centering
    \includegraphics[width=\linewidth]{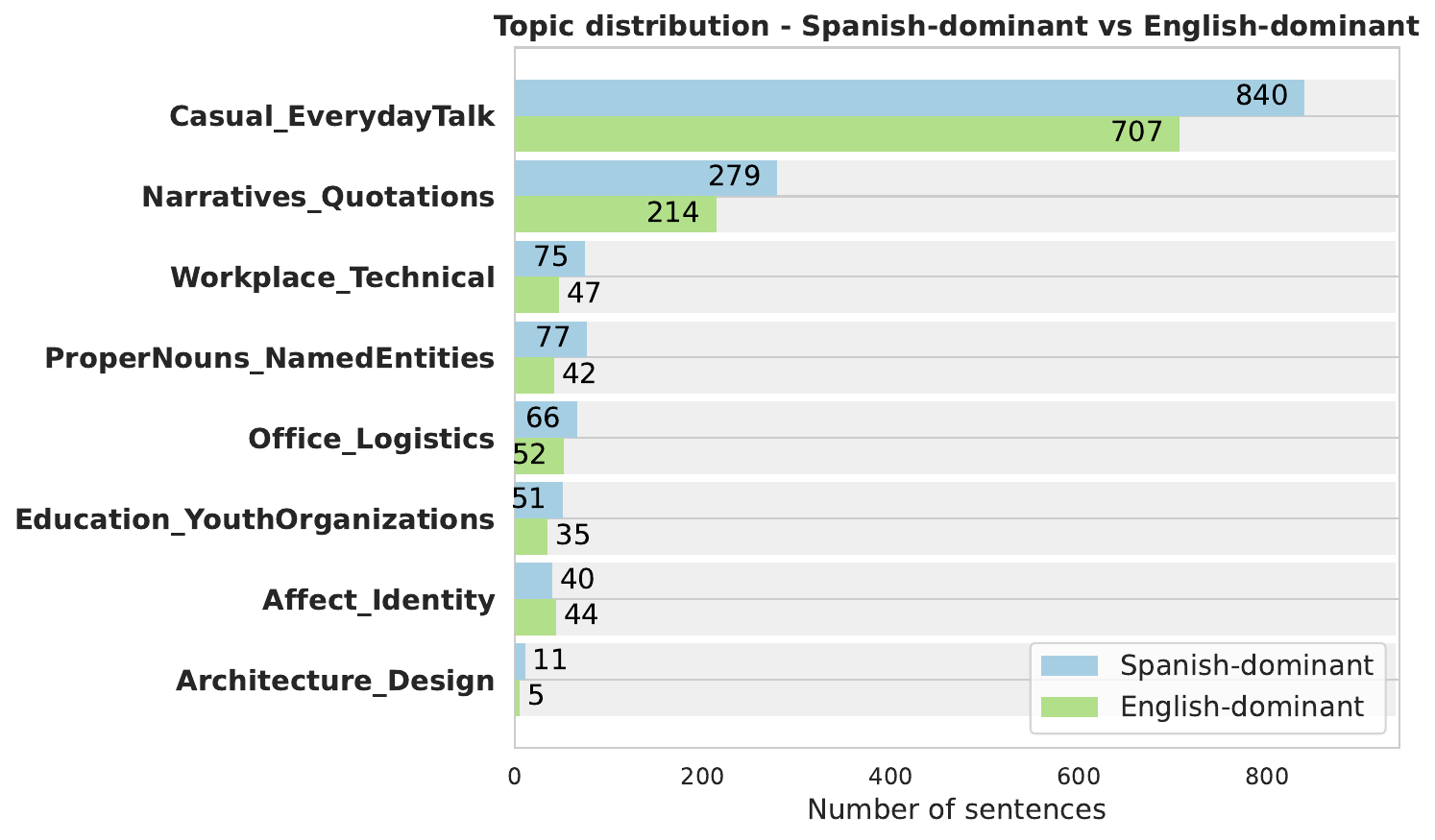}
    \caption{Topic distribution across Spanish- and English-dominant sentences.}
    \label{fig:topic-spa-vs-eng}
  \end{subfigure}

  \begin{subfigure}[t]{\columnwidth}
    \centering
    \includegraphics[width=\linewidth]{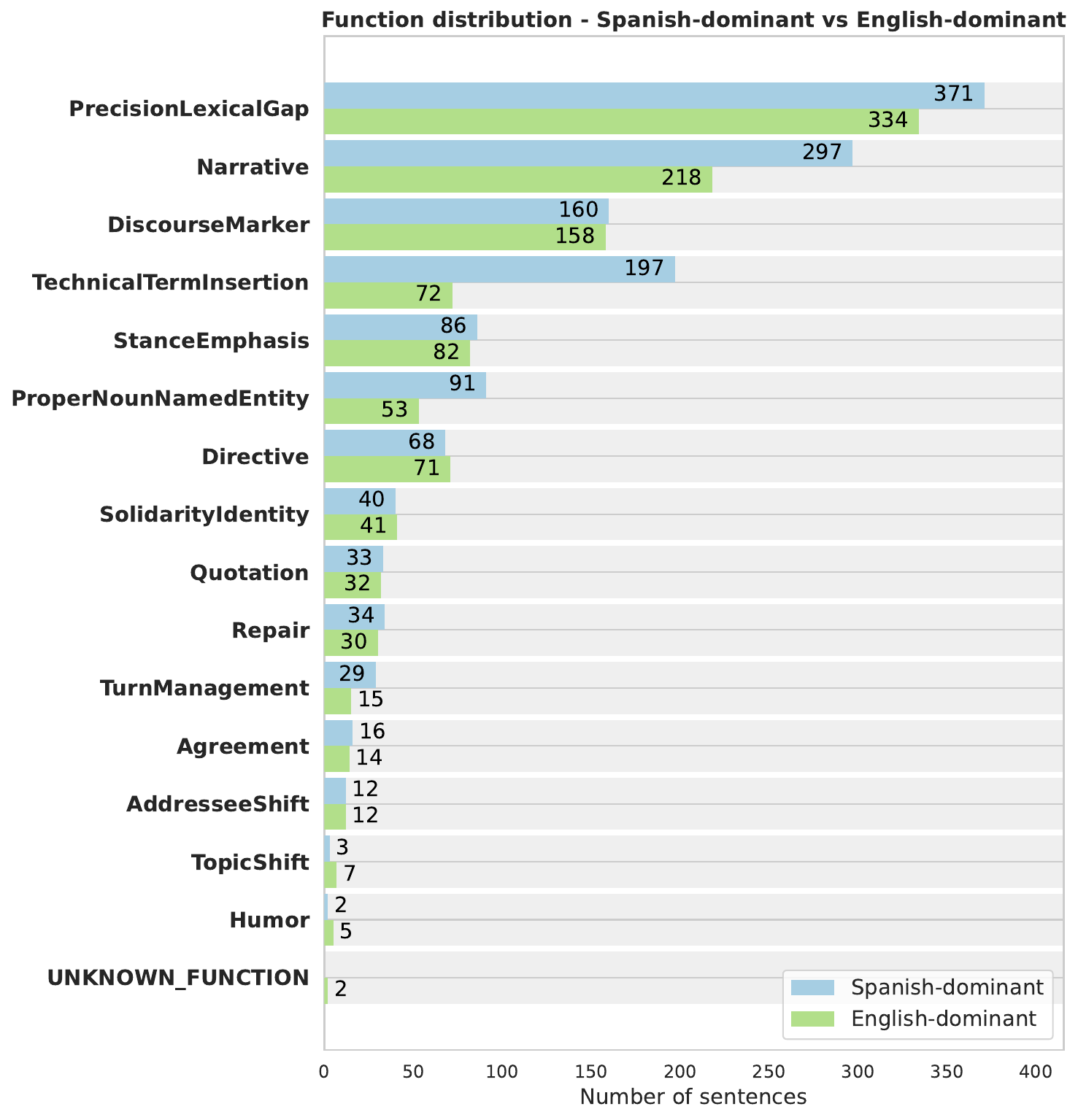}
    \caption{Function distribution across Spanish- and English-dominant sentences.}
    \label{fig:function-spa-vs-eng}
  \end{subfigure}

  \caption{Bilingual asymmetries in the Miami corpus, showing variation in topic and function distributions between Spanish- and English-dominant contexts.}
  \label{fig:miami-dominance}
\end{figure}

Figure~\ref{fig:miami-dominance} illustrates bilingual asymmetry patterns in the Miami corpus. Figure~\ref{fig:topic-spa-vs-eng} compares topic distributions across Spanish- and English-dominant sentences, while Figure~\ref{fig:function-spa-vs-eng}  presents the corresponding discourse functions. Spanish-dominant segments cluster around casual, affective, and narrative domains, reflecting conversational and stance-oriented uses. English-dominant spans show slightly higher frequencies in technical and precision-related categories, whereas both languages are similarly represented in discourse-marker functions. These tendencies align with interactional analyses suggesting that Spanish indexes personal stance and social proximity, while English supports informational precision and referential clarity (\citealt{Bullock2009}; \citealt{Toribio2004}).
Such distributional asymmetries may also be influenced by lexical-level factors that shape bilingual sentence planning. Further evidence for this comes from \citet{Fricke2016}, who found that bilinguals often code-switch in response to lexical triggers such as cognates or shared lexical items. In this view, code-switching operates as a strategy to access otherwise unavailable expressions or to optimize lexical retrieval. Lexical overlap, especially in the non-default or less dominant language, heightens the likelihood of switching, reflecting how bilinguals dynamically manage cross-linguistic resources based on contextual and lexical accessibility.

\subsection{Spanish-Guaraní Corpus}

\subsubsection{Formality-driven topic and genre modelling}
We first inspect how formality (Formal vs Informal) conditions topical and genre distributions in the Spanish-Guaraní dataset. Table~\ref{tab:spa-gua-genres-formality-agg} and Table~\ref{tab:spa-gua-topics-formality-agg} present the same statistics shown as proportions normalized by formality totals (i.e., each formality column is normalized), where categories have been trimmed to the most frequent items (topics: top 15; genres: top 10). These plots highlight register differences (formal institutional uses vs. informal/personal use) across categories. As shown in the above tables, the corpus contains nearly equal proportions of formal and informal sentences; however, their distribution varies substantially by topic. This variation provides a useful basis for examining linguistic phenomena such as code-switching within distinct communicative contexts. Moreover, the observed balance between formal and informal registers appears to correlate with language dominance, suggesting that register and language choice are interdependent dimensions of bilingual discourse.

\begin{table}[t]
\centering
\scriptsize
\setlength{\tabcolsep}{12pt} 
\begin{adjustbox}{max width=\columnwidth}
\begin{tabularx}{\linewidth}{@{}lrrr@{}}
\toprule
\textbf{Genre} & \textbf{Form.(\%)} & \textbf{Inform. (\%)} & \textbf{Total ($n$)} \\
\midrule
News              & 65.1 & 0.3 & 320 \\
Personal          &  0.0 & 72.1 & 271 \\
Politics          & 12.9 & 0.3 & 64  \\
Announcement      & 12.2 & 0.3 & 61  \\
Opinion           &  1.2 & 11.7 & 50  \\
Culture\_Arts     &  5.7 &  2.9 & 39  \\
Entertainment     &  0.0 &  5.9 & 22  \\
Sports            &  0.0 &  2.7 & 10  \\
Others            &  3.8 &  7.2 & 29  \\
\midrule
Total Sentences            & 490  & 376  & 866 \\
\bottomrule
\end{tabularx}
\end{adjustbox}
\caption{Genre formality split (aggregated). Values for all categories after top 8 are summed into the “Others” row. Percent columns are proportions normalized by formality totals.}
\label{tab:spa-gua-genres-formality-agg}
\end{table}

\begin{table}[t]
\centering
\scriptsize
\setlength{\tabcolsep}{5.5pt}
\begin{adjustbox}{max width=\textwidth}
\begin{tabularx}{\linewidth}{@{}lrrr@{}}
\toprule
\textbf{Topic} & \textbf{For. (\%)} & \textbf{Inf. (\%)} & \textbf{Tot. ($n$)} \\
\midrule
UserMention\_Request\_Response     &  0.0 & 30.6 & 115 \\
Humor\_Rant                        &  0.0 & 21.8 &  82 \\
Personal\_Emotional                &  0.0 & 19.9 &  75 \\
Government\_Announcement           & 14.7 &  0.0 &  72 \\
Opinion\_Commentary                &  4.3 &  9.6 &  57 \\
Cultural\_Event\_Festival          &  9.8 &  1.6 &  54 \\
PublicAdministration\_Changes      &  9.6 &  0.3 &  48 \\
Legislation\_Policy                &  7.8 &  0.3 &  39 \\
Corruption\_Donations\_Procurement &  4.7 &  1.3 &  28 \\
Education\_Policy\_University      &  4.1 &  1.1 &  24 \\
Protest\_Report                    &  4.1 &  0.5 &  22 \\
Sports\_Event                      &  1.0 &  4.0 &  20 \\
Crime\_Investigation               &  3.5 &  0.8 &  20 \\
Transport\_PublicSafety            &  3.7 &  0.3 &  19 \\
Legal\_Judicial                    &  3.9 &  0.0 &  19 \\
Indigenous\_CommunityAid           &  3.5 &  0.3 &  18 \\
PublicHealth\_Services             &  3.3 &  0.3 &  17 \\
Infrastructure\_Contract           &  3.5 &  0.0 &  17 \\
Cultural\_Heritage\_Archive        &  2.9 &  0.8 &  17 \\
Others                             &  15.8 & 6.7 & 103 \\
\midrule
Total Sentences            & 490  & 376  & 866 \\
\bottomrule
\end{tabularx}
\end{adjustbox}
\caption{Topic formality split (aggregated). All categories with 15 or less total sentences are combined under “Others.” Percent columns are proportions normalized by formality totals. }
\label{tab:spa-gua-topics-formality-agg}
\end{table}

\begin{figure}[t]
  \centering
  \begin{subfigure}[t]{\linewidth}
    \centering
    \includegraphics[width=\linewidth]
    {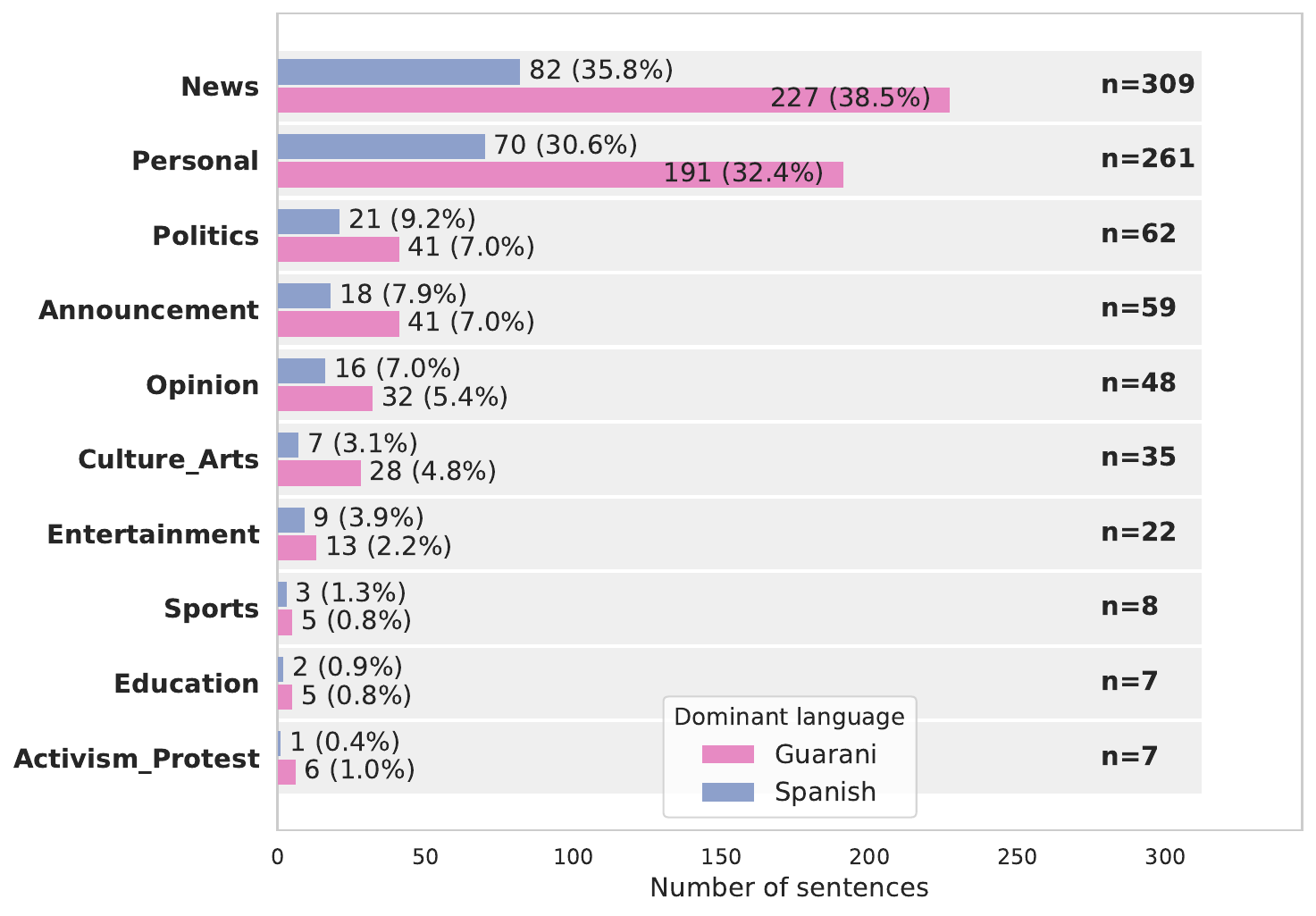}
    \caption{Genre distribution broken down by dominant language (Guaraní-dominant vs Spanish-dominant). Top 10 genres shown.}
    \label{fig:spa-gua-dominance-genre}
  \end{subfigure}

  \begin{subfigure}[t]{\linewidth}
    \centering
    \includegraphics[width=\linewidth]{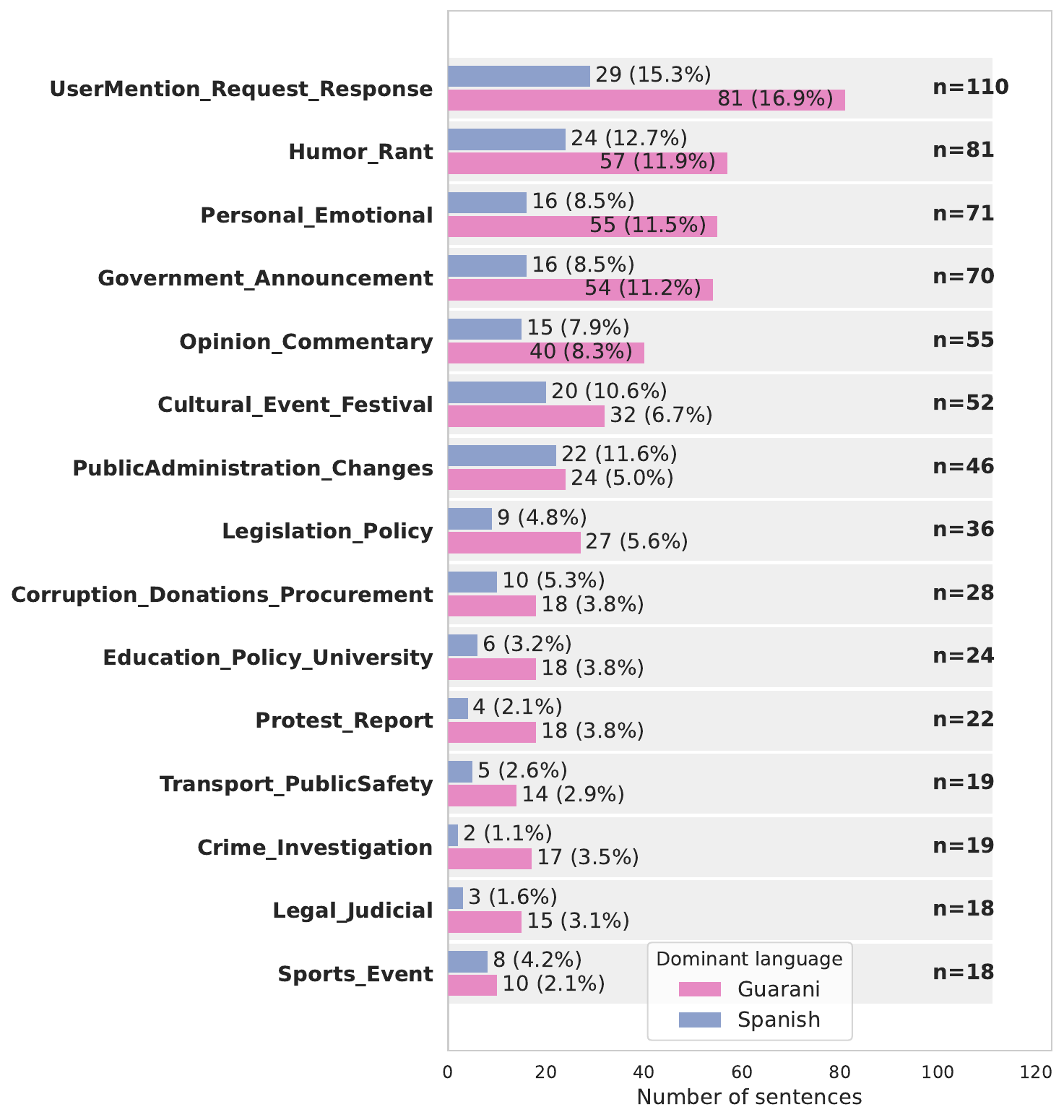}
    \caption{Topic distribution broken down by dominant language (Guaraní-dominant vs Spanish-dominant). Top 15 topics shown.}
    \label{fig:spa-gua-dominance-topic}
  \end{subfigure}
  \caption{Language-dominance comparisons for topics and genres in the Spanish-Guaraní dataset. Each row displays two bars (Guaraní-dominant and Spanish-dominant counts); category lists are trimmed to the most frequent items for clarity.}
  \label{fig:spa-gua-dominance}
\end{figure}

\subsubsection{Language-dominance asymmetries}
Next, we compare category counts across dominant-language splits (Guaraní-dominant vs Spanish-dominant). Figure~\ref{fig:spa-gua-dominance-topic} and Figure~\ref{fig:spa-gua-dominance-genre} present category counts with two grouped bars per category (one bar for each dominant-language class). These plots reveal which topics and genres are more commonly associated with Guaraní- vs Spanish-dominant sentences. Guaraní-dominant texts are concentrated in \textit{"Government\_Announcement"}, \textit{"PublicAdministration\_Changes"}, and \textit{"Indigenous\_CommunityAid"}, reflecting the language’s institutional and communal authority. Spanish-dominant texts emphasize \textit{"Personal\_Emotional"}, \textit{"Humor\_Rant"}, and \textit{"UserMention\_Request\_Response"}, highlighting interpersonal and expressive use. The resulting division between formal Guaraní and informal Spanish supports long-standing observations of diglossic role distribution in Paraguay (\citealt{Rubin1968}; \citealt{Gynan2001}; \citealt{Zajicova2019}) but now emerges from corpus-scale quantitative evidence.

\section{Discussions}

This study demonstrates that LLMs can serve as practical tools for enriching bilingual corpora with topic and sociolinguistic annotations. By combining GPT-based inference with interpretable category schemas, we achieved high annotation accuracy across both high- and low-resource code-switched datasets. Beyond methodological validation, the analysis revealed distinct community-level dynamics: Miami speakers showed sharper gender- and language-dominant contrasts, whereas the Spanish-Guaraní community displayed complementary language use across registers, with Guaraní favored in formal discourse and Spanish in informal, affective interaction.

\paragraph{Methodological and Resource Implications.}
The proposed  pipeline provides a scalable and interpretable approach for semi-automatic corpus enrichment. By leveraging structured instructions and few-shot examples, the method reduces annotation cost while maintaining linguistic transparency. The resulting topic- and function-annotated corpora expand the empirical base for analyzing bilingual discourse and improve the accessibility of training resources for multilingual NLP, particularly in underrepresented languages. Future refinements could integrate adaptive prompt optimization to assess and mitigate potential contextual biases in LLM-generated labels.

\paragraph{Sociolinguistic and Theoretical Insights.}
The integration of sociolinguistic metadata with discourse-pragmatic annotation provides a quantitative basis for examining how language choice reflects social and cognitive constraints. Gender- and dominance-based asymmetries in the Miami data, alongside register-based differentiation in Spanish-Guaraní, align with sociolinguistic theories of stance, identity, and alignment (\citealt{Poplack1980}; \citealt{Toribio2004}). These findings illustrate that code-switching functions as a socially strategic resource rather than linguistic noise. At the lexical level, our results also resonate with psycholinguistic evidence that lexical accessibility and cognate activation influence switch likelihood (\citealt{Fricke2016}). Future modeling could operationalize these mechanisms by incorporating measures of lexical overlap and semantic similarity, thereby linking discourse-level switching patterns with cognitive processes of bilingual word retrieval.

\paragraph{Advancing Topic Modeling for Bilingual Data.}
Current annotation schemes rely on fixed topic taxonomies that promote comparability but may constrain discovery. Dynamic topic generation by LLMs offers a way to capture fluid, context-dependent thematic structure in spontaneous discourse. Because topics in natural interaction frequently overlap, such as politics intersecting with education or identity, proto-ablation-based and embedding-based representations may better capture semantic proximity and topic coherence (\citealt{bianchi-etal-2021-pre}). Incorporating these representations can disambiguate semantically adjacent categories and yield richer, more flexible bilingual topic models.

\paragraph{Toward Integrated Multilevel Modeling.}
The next step is to integrate the annotation of discourse and pragmatics with syntactic, semantic, and sociolinguistic analyzes to construct a multilevel model of bilingual production. Linking topic and function annotations with grammatical and dependency structures will allow researchers to trace how discourse roles interact with structural switch points and grammatical constraints. Such integration can connect item-level accessibility with discourse-level planning, offering a cognitively informed and computationally tractable account of how bilingual speakers manage alignment and coherence across typologically distinct languages. Ultimately, this approach points toward a unified framework for modeling social bilingual discourse across a variety of multilingual communities.

\section{Conclusion}

This paper introduced an LLM-assisted annotation framework for topic and sociolinguistic labeling of bilingual corpora, evaluated on Spanish-English and Spanish-Guaraní datasets. The method achieved high annotation reliability while maintaining interpretability, enabling scalable enrichment of code-switched data. The resulting resources and framework contribute to expanding the empirical and typological scope of bilingual discourse research, particularly for underrepresented language pairs. Beyond the specific corpora released, this work demonstrates how LLMs can serve as practical instruments for linguistically grounded resource development. Future research will extend this framework to integrate syntactic and affective layers, such as dependency relations, sentiment, and stance, toward building comprehensive, socio-computational models of code-switching.

\section{Acknowledgements}
This work was supported by the School of International Letters and Cultures (SILC) at Arizona State University. Special thanks go to the Universidad Nacional de Asunción and especially to Eliodora Verón and Ricardo Peloso for their annotation reviews and contributions. We also thank the organizers of IberLEF~2023 for access to the GUA-SPA dataset.

\section{Ethical Considerations}

All data used are publicly available or anonymized (Bangor Miami Corpus, IberLEF GUA-SPA). No personally identifiable information was processed. We acknowledge potential cultural bias in LLM outputs and have incorporated manual verification to ensure representational fairness, especially for Indigenous languages. All examples respect community norms and licensing agreements.


\section{Bibliographical References}
\bibliographystyle{lrec2026-natbib}
\bibliography{lrec2026}
\section{Language Resource References}
\label{lr:ref}
\bibliographystylelanguageresource{lrec2026-natbib}
\bibliographylanguageresource{languageresource}

\end{document}